\pdfoutput=1

\documentclass[11pt]{article}

\usepackage{acl}

\usepackage{times, xspace}
\usepackage{latexsym}
\newcommand{\our}{\text{DSRE-NLI}\xspace}
\usepackage{amsmath, amsfonts}
\DeclareMathOperator*{\argmax}{arg\,max}
\usepackage{amssymb}
\usepackage{xcolor}
\usepackage{multirow}
\usepackage{graphicx}
\usepackage{lscape}
\usepackage{pifont}
\usepackage{balance}
\newcommand{\cmark}{\ding{51}}%
\newcommand{\xmark}{\ding{55}}%

\usepackage[ruled, linesnumbered, vlined]{algorithm2e}
\usepackage[T1]{fontenc}

\usepackage[utf8]{inputenc}

\usepackage{microtype}

\title{Improving Distantly Supervised Relation Extraction by \\ Natural Language Inference}

\author{Kang Zhou,\hspace{2mm} Qiao Qiao,\hspace{2mm} Yuepei Li,\hspace{2mm} Qi Li \\
  Department of Computer Science, Iowa State University, Ames, Iowa, USA \\
  \texttt{\{kangzhou, qqiao1, liyp0095, qli\}@iastate.edu} \\}
  
\begin{document}
\maketitle
\begin{abstract}
To reduce human annotations for relation extraction (RE) tasks, distantly supervised approaches have been proposed, while struggling with low performance. In this work, we propose a novel \our framework, which considers both distant supervision from existing knowledge bases and indirect supervision from pretrained language models for other tasks. \our energizes an off-the-shelf natural language inference (NLI) engine with a semi-automatic relation verbalization (SARV) mechanism to provide indirect supervision and further consolidates the distant annotations to benefit multi-classification RE models. The NLI-based indirect supervision acquires only one relation verbalization template from humans as a semantically general template for each relationship, and then the template set is enriched by high-quality textual patterns automatically mined from the distantly annotated corpus. With two simple and effective data consolidation strategies, the quality of training data is substantially improved. Extensive experiments demonstrate that the proposed framework significantly improves the SOTA performance (up to 7.73\% of F1) on distantly supervised RE benchmark datasets.

\end{abstract}

\section{Introduction}\label{sec:intro}
Relation extraction~(RE) has been studied intensively in the past years~\cite{zhou2021improved, zhang2017position, zelenko2003kernel}. It aims to extract the relations among entities from text and plays an important role in various natural language processing~(NLP) tasks such as knowledge graph construction~\cite{distiawan2019neural}, question answering~\cite{yu2017improved}, and text summarization~\cite{hachey2009multi}. In this paper, we define the RE task as to identify the pre-defined relation for a pair of entity mentions in a given sentence.

\begin{figure}[t]
    \centering
    \includegraphics[width=0.48\textwidth]{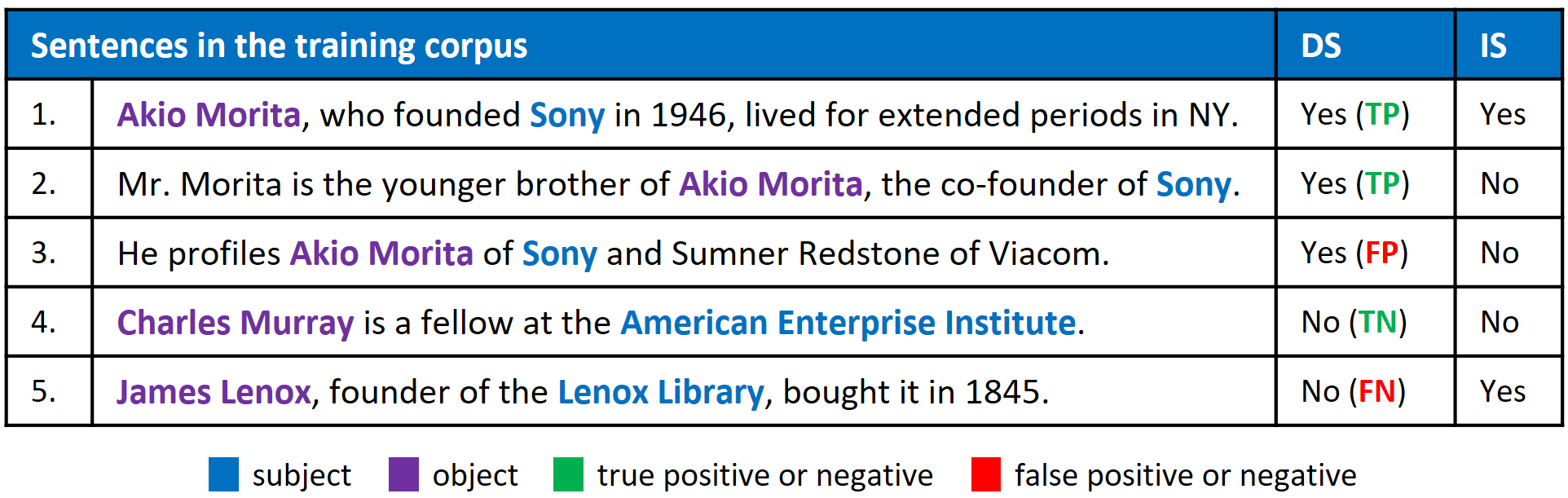}
    \caption{\small Annotation examples of distant supervision (DS) and indirect supervision (IS) for relation \texttt{founders}.}
    \label{fig:examples}
    \vspace{-5mm}
\end{figure}

Due to the cost of human annotations for RE tasks, researchers have been trying to develop alternative approaches without requiring human annotations. Specially, two practically appealing learning strategies have shown promising results: distantly supervised learning using large noisy training data and zero-shot learning using indirect supervision. 

Distant supervision acquires massive annotations using existing in-domain knowledge bases~\cite{ma2021sent, zheng2019diag, jia2019arnor}. 
A commonly adopted distant annotation process for RE tasks is that if two entities participate in a relation in the knowledge base, then all sentences in the training corpus with these two entities are labeled as positive examples of that relation \cite{mintz2009distant, riedel2010modeling}. 

Distantly supervised methods usually face high label noise in training data due to the annotation process, since not all sentences with the entity pair express the relationship. Figure \ref{fig:examples} shows some examples of distant annotations for the \texttt{founders} relationship. Since \texttt{Akio Morita} and \texttt{Sony} have the \texttt{founders} relationship in the knowledge base, all sentences with the two entities are labeled as positive, introducing the false positive problem. On the other hand, due to the limited coverage of the knowledge base, Sentence 5 is labeled as negative, introducing the false negative problem. Therefore, existing distantly supervised methods are proposed to tackle the noise in training data \cite{ma2021sent, jia2019arnor, zheng2019diag, lin2016neural, zeng2015distant}. 

Recently, indirectly supervised methods have taken advantage of pretrained models for other NLP tasks to solve RE tasks in the zero-shot setting. For example, RE tasks have been reformulated as question answering problems~\cite{levy2017zero}, as natural language inference (NLI) tasks~\cite{obamuyide2018zero, sainz2021label}, and as text summarization tasks~\cite{lu2022summarization}. Performance in the zero-shot setting, however, still has a significant gap, and highly relies on the quality and diversity of relation paraphrase templates. For example, we apply the NLI-based method \cite{sainz2021label} with a relation verbalization template \texttt{\{subj\} was founded by \{obj\}} for sentences in Figure \ref{fig:examples}. For Sentence 2, the model cannot align \texttt{co-founder} meaning with it and thus mislabels. 

In this work, we introduce indirect supervision into distantly supervised RE (DSRE) tasks for the first time to improve its performance. 
Specifically, the proposed \our energizes an off-the-shelf NLI engine with a novel semi-automatic relation verbalization (SARV) mechanism to diagnose label noise in distant annotations. To involve as little as possible human effort in the relation verbalization process, we acquire only one semantically general template from humans for each relationship. To improve the semantic diversity of relation templates, we conduct an NLI involved textual pattern mining and grouping process to enrich the template set of each relation by choosing high-quality textual patterns from the distantly annotated corpus. These relation verbalization templates are used as-is for an NLI-based zero-shot RE on the training corpus to provide indirect supervision, which further consolidates the distant annotations with two simple and effective strategies. Finally, the consolidated training data are used to train traditional multi-class RE models for prediction.


In empirical studies, we use two real DSRE benchmark datasets and a simulated dataset to evaluate the proposed method. The results show that the proposed \our consistently outperforms the state-of-the-art models by a large margin. The simulation and ablation studies show that \our can create a high-quality training data.

\section{Related Work}\label{sec:related}

\paragraph{Distantly Supervised Relation Extraction.} 

            
\citet{mintz2009distant} propose the DSRE task for the first time. It assumes that if two entities participate in a relation, then all sentences mentioning the two entities express that relation. \citet{riedel2010modeling} argue that the assumption is too strong in real practice, so they modify the assumption as if two entities participate in a relation, at least one sentence that mentions the two entities express that relation. This assumption is further modified to allow multiple labels for an entity pair \cite{hoffmann2011knowledge, surdeanu2012multi}. 

The research focus of distantly supervised methods is to tackle the noise in training data, especially the false positive problem. One strategy is to apply the multi-instance learning framework~\cite{hoffmann2011knowledge, surdeanu2012multi, zeng2015distant, lin2016neural, jiang2018revisiting}. These methods form positive bags of sentences for a relation following the expressed-at-least-once assumption. Then the learner is trained on sentence bags instead of individual sentences. However, \citet{feng2018reinforcement} first report that bag-level methods struggle in sentence-level prediction. 

Another line of approaches adopts various sentence-level denoising strategies. For example, pattern mining methods have been shown to be effective in reducing false positives in training data~\cite{li2018truepie,qu2018weakly, zheng2019diag, jia2019arnor}. However, pattern based methods tend to have low recall since pattern matching is a restricted process.
Some methods apply reinforcement learning to automatically recognize false positive samples \cite{feng2018reinforcement, qin2018robust,he2020improving}. \citet{ma2021sent} employ negative training to denoise and use a relabeling mechanism to iteratively train RE models.

\paragraph{Zero-Shot Relation Extraction.}
Zero-shot RE methods apply indirect supervision and convert RE tasks to other NLP tasks \cite{levy2017zero,obamuyide2018zero, sainz2021label, lu2022summarization}. For the zero-shot setting, human annotators do not label any samples, but are usually asked to generate relation paraphrase templates. For example, \citet{levy2017zero} use crowd-sourcing to generate question templates. \citet{sainz2021label} ask human annotators to generate verbalization templates for at most 15 minutes per relation. These methods show that the few-shot setting performs significantly better than the zero-shot setting, but requires some human efforts for annotating training samples.

In this paper, we adopt NLI-based indirect supervision for two reasons: 1) relation verbalization templates can be directly obtained from textual patterns mined from distantly labeled corpus; 2) the inference step from NLI to RE is straightforward. 
\section{Preliminary} \label{sec:pre}

\subsection{RE Task Definition}
We formalize the RE task as follows. Let $\boldsymbol{x} = [x_1, ..., x_n]$ denote a sentence, where $x_i$ is the $i$-th token. An entity pair $(\boldsymbol{e}_{subj}, \boldsymbol{e}_{obj})$, referring to the subject and object entities, respectively, is identified in the sentence, where $\boldsymbol{e}_{subj} = [x_{ss}, ..., x_{se}]$ and $\boldsymbol{e}_{obj} = [x_{os}, ..., x_{oe}]$ are two non-overlapping consecutive spans. Given an instance, which includes the sentence $\boldsymbol{x}$, and the specific positions and entity types of $\boldsymbol{e}_{subj}$ and $\boldsymbol{e}_{obj}$, the goal is to predict the relation $r \in \mathcal{R} \cup \{\rm NA\}$ that holds between $\boldsymbol{e}_{subj}$ and $\boldsymbol{e}_{obj}$, where $\mathcal{R}$ is a pre-defined relation set, and $\rm NA$ indicates that no relation from $\mathcal{R}$ is expressed between them.

\subsection{Distant Annotation} \label{sec:DA}
In order to construct distantly annotated training data, named entity mentions are first recognized from the corpus by named entity recognition (NER) methods. Then, by exact string matching, the named entity mentions are linked to the entities in the knowledge base (e.g., Freebase) that covers the relations of interest (i.e., $\mathcal{R}$). If a sentence contains two entity mentions that have a relation of interest in the knowledge base, then a corresponding instance will be generated and labeled as the relation type $r$. Otherwise, an instance labeled as NA will be generated. 


\subsection{NLI for RE}\label{sec:NLI}
Since our framework will utilize the reformulation of RE into NLI to obtain indirect supervision, here we briefly introduce the reformulation schema proposed in \citet{sainz2021label}. Given a textual premise and a hypothesis, NLI, also known as textual entailment, is to determine whether the premise entails, or is neutral to, or contradicts the hypothesis. The reformulation requires three sub-processes: relation verbalization, NLI input generation, and relation inference. 

Relation verbalization is to verbalize a relation by a simple paraphrase of only a few tokens. Such paraphrases are called verbalization templates. For example, \texttt{\{subj\} was founded by \{obj\}} is a verbalization template for the relation \texttt{/business/company/founders} from Freebase, where \texttt{\{subj\}} and \texttt{\{obj\}} are placeholders. Note that a relation can have multiple verbalization templates. For example, \texttt{\{obj\} is a founder of \{subj\}} can be another template for \texttt{founders} relation. For a relation $r \in \mathcal{R}$, \citet{sainz2021label} give 15 minutes to human annotators to generate several templates and construct a verbalization template set $\mathcal{T}_r$, and $\lvert \mathcal{T}_r \rvert \geq 1$.

NLI input generation is to generate premise-hypothesis pairs for each sentence. These pairs will be taken as input by an NLI model. Given a template $\boldsymbol{t} \in \mathcal{T}_r$ and a sentence $\boldsymbol{x}$ mentioning two entities $\boldsymbol{e}_{subj}$ and $\boldsymbol{e}_{obj}$, NLI input generation yields a premise-hypothesis pair ($\boldsymbol{x}, \boldsymbol{h}$), where $\boldsymbol{h} =  hyp(\boldsymbol{t}, \boldsymbol{e}_{subj}, \boldsymbol{e}_{obj})$ with $hyp(\cdot)$ substituting placeholders in the template with actual entities. 


Relation inference is to infer the relationship expressed by two entity mentions in a sentence from the outputs of an NLI model. Taking a premise-hypothesis pair as input, an NLI model with a softmax output layer yields a probability distribution $\boldsymbol{P}_{NLI}(\boldsymbol{x}, \boldsymbol{h})$ over entailment (E), neutrality (N), and contradiction (C). Then, the probability of ($\boldsymbol{e}_{subj}, \boldsymbol{e}_{obj}$) expressing relation $r$ in sentence $\boldsymbol{x}$ is computed by:
\begin{align*}
     P_r(\boldsymbol{x}, \boldsymbol{e}_{subj}, \boldsymbol{e}_{obj}) =  \delta_r\max_{\boldsymbol{t} \in \mathcal{T}_r}P_{NLI}^{E}(\boldsymbol{x}, \boldsymbol{h}),
\end{align*}
where $P_{NLI}^{E}$ is the probability of entailment and $\delta_r$ is an indicator function to tackle entity type constraints. Occasionally, a template can verbalize more than one relation, which will bring ambiguity in later inference. For example, \texttt{\{subj\} was born in \{obj\}} can verbalize both \texttt{country\_of\_birth} and \texttt{city\_of\_birth}. $\delta_r$ considering NER type information can tackle this issue. $\delta_r(\boldsymbol{e}_{subj}, \boldsymbol{e}_{obj}) = 1$ if ${\rm NER}(\boldsymbol{e}_{subj}, \boldsymbol{e}_{obj})\in \mathcal{E}_r$, otherwise 0, where $\mathcal{E}_r$ is a set of NER type constraints for relation $r$. For example, the former requires ${\rm NER}(\cdot)\in \{\texttt{PERSON:COUNTRY}\}$, while the latter requires ${\rm NER}(\cdot)\in \{\texttt{PERSON:CITY}\}$.
Then, the final predicted relation $\hat{r}$ of ($\boldsymbol{e}_{subj}, \boldsymbol{e}_{obj}$) in $\boldsymbol{x}$ is given by:
\begin{align*}
    \hat{r} = \argmax_{r \in \mathcal{R}} P_r(\boldsymbol{x}, \boldsymbol{e}_{subj}, \boldsymbol{e}_{obj}).
\end{align*}
Note that if $\max_{r \in \mathcal{R}} P_r(\boldsymbol{x}, \boldsymbol{e}_{subj}, \boldsymbol{e}_{obj}) < \tau$, then $\hat{r} = \rm NA$, where $\tau$ is a threshold that is a hyperparameter. 

\citet{sainz2021label} propose two settings: zero- and few-shot. Zero-shot directly uses a pretrained NLI model, while few-shot uses training data with ground truth labels to fine-tune the pretrained NLI model. Zero-shot shows a significant gap in performance comparing with few-shot.

\section{Methodology}\label{sec:method}
\begin{figure}[t]
    \centering
    \includegraphics[width=0.48\textwidth]{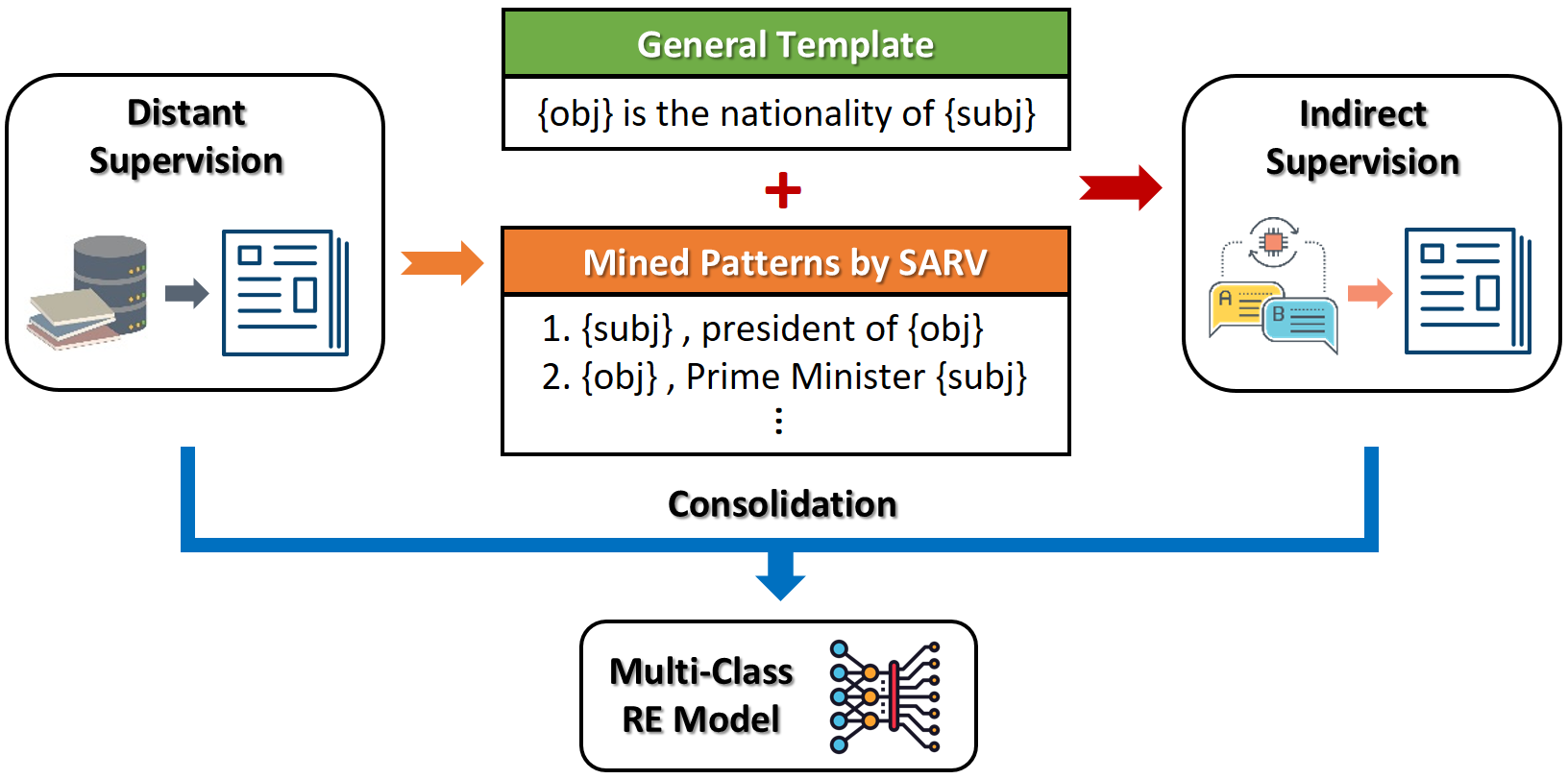}
    \caption{\small DSRE-NLI framework. Template and pattern examples are from relation \texttt{nationality}.}
    \label{fig:framework}
    \vspace{-3mm}
\end{figure}

This section introduces the proposed framework \our. Figure~\ref{fig:framework} illustrates the architecture. We use the distant annotation process in Section~\ref{sec:DA} to provide distant supervision (DS) and the NLI-based zero-shot RE in Section~\ref{sec:NLI} to provide indirect supervision (IS). The textual patterns are mined from distantly annotated corpus. The candidate patterns are then presented to human annotators. The selected patterns with one human generated general template are used for the NLI model. The distant annotations are consolidated with results of the NLI model and used for the multi-class RE model training. 


\subsection{Semi-Automatic Relation Verbalization} 

The performance of the NLI-based zero-shot RE is highly dependent on the quality and diversity of verbalization templates, where template quality reflects if a template accurately conveys the semantic meaning of the relationship, and template diversity reflects the semantic coverage of templates. As introduced in Section \ref{sec:pre}, the reformulation of RE into NLI requires human generated templates.

However, human written templates are specifically prone to semantic generalization and duplication, and may not fit the writing style of the specific corpus, leading to poor performance and computational inefficiency in the zero-shot setting. 
Further more, a pretrained NLI model determines text entailment differently from humans, making it harder for humans to propose diverse useful templates. For example, an annotator may provide both \texttt{\{subj\} is the parent of \{obj\}} and \texttt{\{obj\} is the son of \{subj\}} for relation \texttt{children}, believing the two templates increase the semantic diversity than a single one. However, a pretrained NLI model may determine that the latter actually strongly entails the former, indicating that the latter has no contribution to the diversity because of an observed transition property that if sentence $\boldsymbol{x}$ entails template $\boldsymbol{t}_1$, and $\boldsymbol{t}_1$ entails template $\boldsymbol{t}_2$, then $\boldsymbol{x}$ entails $\boldsymbol{t}_2$. On the other hand, an annotator may think \texttt{\{subj\} was founded by \{obj\}} is similar to \texttt{\{obj\} is a co-founder of \{subj\}} for relation \texttt{founders} and thus omit one. However, a pretrained NLI model may not determine that they strongly entail each other, which means that the two templates may improve the semantic coverage. Therefore, without prior knowledge of the corpus and the NLI model behavior, it is hard for an annotator to propose proper and diverse templates for specific relation types.

To overcome the template fitness and diversity issues, we propose a novel semi-automatic relation verbalization (SARV) method by considering the content of the distantly annotated corpus, so that an annotator can efficiently select templates of higher quality and higher diversity for each relationship. 



We propose to generate template candidates by textual pattern mining and grouping. Pattern mining from the distantly annotated corpus can discover various expression styles in the specific corpus. Pattern grouping can discard semantically duplicated templates, and thus improve computational efficiency. Moreover, pattern grouping can accumulate the pattern frequency in individual groups so as to highlight truly useful semantic patterns.

\subsubsection{Pattern Mining and Grouping}
For each relation type, we collect all instances and conduct a simple pattern mining that directly takes the token sequence between subject and object entities, which is easy and efficient. To ensure the quality of candidate patterns, we use three criteria to filter patterns: 1) pattern frequency should be higher than a threshold; 2) pattern length should be shorter than a threshold; 3) a pattern must contain at least one non-stop-word.
Advanced pattern mining techniques can also be applied, such as PATTY \cite{nakashole2012patty}, mining patterns from shortest dependency path between a pair of entities, and MetaPAD \cite{jiang2017metapad}, mining patterns from entire sentences. 

Previous pattern mining methods use shallow features to group patterns into semantic clusters, such as pattern lexicon and extraction overlaps \cite{li2018pattern, nakashole2012patty,jiang2017metapad} and pattern embeddings \cite{li2018truepie}. These methods do not group patterns with semantic understanding of the patterns. Therefore, we propose an NLI-based pattern grouping method.

We first define semantic duplication from the perspective of the NLI task as follows.

\noindent \textit{\textbf{Definition}.}
Given patterns $\boldsymbol{p}_1$ and $\boldsymbol{p}_2$ of relation $r$, and $\lvert \boldsymbol{p}_1 \rvert \geq \lvert \boldsymbol{p}_2 \rvert$, if $\boldsymbol{p}_1$ strongly entails $\boldsymbol{p}_2$ (i.e., $P_{NLI}^{E}(\boldsymbol{p}_1, \boldsymbol{p}_2) \geq \tau$) by an NLI model, then we say $\boldsymbol{p}_1$ is semantically duplicated to $\boldsymbol{p}_2$.

Pattern grouping aims to reduce the semantic duplication and only maintain the semantically distinct representative patterns for each relation. To do so, we first rank the initial patterns from the mining results by their length. Then starting from the shortest pattern, we recursively use a pretrained NLI model to determine which patterns from the rest longer ones are semantically duplicated to the pattern.
If $\lvert \boldsymbol{p}_j \rvert \geq \lvert \boldsymbol{p}_i \rvert$ and $\boldsymbol{p}_j$ is duplicated to $\boldsymbol{p}_i$, then $\boldsymbol{p}_j$ is grouped with $\boldsymbol{p}_i$, and $\boldsymbol{p}_j$ will not participate in the subsequent grouping process. After the grouping process, the shortest (semantically dense) patterns of individual groups will be the final representative patterns for the specific relation, and they will be ranked by their group frequencies (i.e., accumulated frequencies of all patterns in the group) for the further manual screening.

\subsubsection{Template Generation and Selection}
Since a pre-defined relation label usually conveys the semantic meaning of the relation, an annotator can easily propose one semantically general verbalization template as a general template for each relation even without much domain knowledge. This can guarantee there is at least one template of high quality. It is important especially for long-tail relation types where the patterns are sparse. 
With the general template, we can continue shrinking the pattern candidates of each relation by removing patterns that are semantically duplicated to it. From the results, human annotators further select high-quality patterns as additional templates. 

Algorithm \ref{alg:pg} summarizes the SARV process, where `$\Rightarrow$' denotes `semantically duplicated to', and $f_{\boldsymbol{p}}$ denotes the frequency of $\boldsymbol{p}$. Line 1 is conducting pattern mining. Lines 2-10 are conducting pattern grouping. Lines 11-16 are performing template generation and selection.

\SetKwComment{Comment}{/* }{ */}
\SetKwInput{KwData}{Input}
\SetKwInput{KwResult}{Output}

\begin{algorithm}[t]
\small
\caption{SARV}\label{alg:pg}
\KwData{Distantly labeled corpus $\mathcal{C}$ and relation label $r$}
\KwResult{Template set $\mathcal{T}_r$}
mine an initial pattern set $\mathcal{P}_r^{initial}$ from $\mathcal{C}$\;
sort $\mathcal{P}_r^{initial}$ by pattern length $\lvert \boldsymbol{p} \rvert$ in increasing order\;
$\mathcal{P}' \gets \emptyset$\;
\For{$\boldsymbol{p}_i$ \rm{\textbf{in}} $\mathcal{P}_r^{initial}$}{
    $\mathcal{P}' \gets \mathcal{P}' \cup \{\boldsymbol{p}_i\}$\;
    \For{$\boldsymbol{p}_j$ \rm{\textbf{in}} $\mathcal{P}_r^{initial} - \mathcal{P}'$}{
        \If{$f_{\boldsymbol{p}_j} > 0$ \rm{\textbf{and}} $\boldsymbol{p}_j \Rightarrow \boldsymbol{p}_i$}{
            $f_{\boldsymbol{p}_i} \gets f_{\boldsymbol{p}_i} + f_{\boldsymbol{p}_j}$\;
            $f_{\boldsymbol{p}_j} \gets 0$\;
        }
    }
}
$\mathcal{P}_r^{grouped} \gets \mathcal{P}_r^{initial} - \{\boldsymbol{p} \mid f_{\boldsymbol{p}} = 0\}$\;
request a general template $\boldsymbol{t}_r$, and add into $\mathcal{T}_r$\;
\For{$\boldsymbol{p}_i$ \rm{\textbf{in}} $\mathcal{P}_r^{grouped}$}{
    \If{$\boldsymbol{p}_i \Rightarrow \boldsymbol{t}_r$}{
        $f_{\boldsymbol{p}_i} \gets 0$\;
    }
}
$\mathcal{P}_r^{grouped} \gets \mathcal{P}_r^{grouped} - \{\boldsymbol{p} \mid f_{\boldsymbol{p}} = 0\}$\;
sort $\mathcal{P}_r^{grouped}$ by pattern frequency, then select high-quality patterns from $\mathcal{P}_r^{grouped}$, and add into $\mathcal{T}_r$\;
\Return $\mathcal{T}_r$\;
\end{algorithm}

\subsection{Training Data Consolidation}
Recall from Section \ref{sec:DA}, the distant supervision can annotate a set of instances for each relation $r \in \mathcal{R} \cup \{\rm NA\}$. Our first consolidation strategy is to use a pretrained NLI model to filter all distantly annotated sets. For the set of instances annotated as $r$ by distant supervision, if an instance is not predicted as $r$ by the NLI model,
it will be removed.
That is, the intersection of the results from DS and IS. We call this strategy as IPIN (\textbf{I}ntersection of \textbf{P}ositives and \textbf{I}ntersection of \textbf{N}egatives).

Another strategy is to use only IS to construct the set of instances for relation $r \in \mathcal{R}$ and use the intersection of DS and IS to construct the set of instances for $\rm NA$. The reason is that the intersection of positive instances may reduce the size of positive instances significantly and impact the learning effect especially for long-tail relation types, and IS (i.e., the NLI model) can generally provide predictions with high precision. For $\rm NA$, since both DS and IS can produce large sets, the intersection of sets can filter out false negatives and still maintain sufficient instances. We call this strategy NPIN (\textbf{N}LI \textbf{P}ositives and \textbf{I}ntersection of \textbf{N}egatives).

\subsection{Multi-Class RE Model} \label{sec:MRE}

Using the consolidated training data, we can train a multi-class RE model. We adopt the architecture proposed by \citet{zhou2021improved} using entity mask as the entity representation strategy. Although in their experiments, the typed entity marker strategy performed better, for the DSRE task, we find that entity mask is more tolerant to the label noise. This is because that the pretrained language model (BERT in our case) has strong prior knowledge about entities and the noise in training can strengthen the reliance on prior knowledge. Without the surface names of entities, the model learns from the context of the sentences instead, and thus more robust to label noise.
\section{Experiments}\label{sec:exp}
In this section, we evaluate \our framework and compare with other baseline methods.

\subsection{Experiments on Real Distantly Annotated Datasets}\label{sec:realexp}

\subsubsection{Datasets and Evaluation Metrics}
We conduct experiments to evaluate the proposed framework on the widely-used public dataset: New York Times (NYT), which is a large-scale distantly labeled dataset constructed from NYT corpus using Freebase as the distant supervision \cite{riedel2010modeling}. Recently, \citet{jia2019arnor} manually labeled a subset of the data for a more accurate evaluation and constructed two versions of the dataset: NYT1 and NYT2\footnote{\url{ https://github.com/PaddlePaddle/Research/tree/48408392e152ffb2a09ce0a52334453e9c08b082/NLP/ACL2019-ARNOR}}. The latter version is released after their paper publication. The statistics of the two datasets are summarized in Table~\ref{table:dataset}, and more details about the instance generation can be found in Appendix Section \ref{apx:inst}. We report the evaluation results in terms of precision, recall and F1 score. For all metrics, the higher the better.

\begin{table}[t]
\centering
\resizebox{0.45\textwidth}{!}{%
\begin{tabular}{cl|ccc}
\hline
\multicolumn{2}{c}{Dataset}               & NYT1 & NYT2  & TACREV\\\hline
\multicolumn{2}{c|}{\# relation}          & 10      & 11    & 41\\\hline
\multirow{2}{*}{Train} & \# total   inst & 376,355 & 373,643 & 68,124\\
                       & \# pos inst     & 95,519  & 92,807  & 13,012\\\hline
\multirow{2}{*}{Dev}   & \# total   inst & 2,379   & 4,569   & 22,631\\
                       & \# pos inst     & 338     & 973     & 5,300\\\hline
\multirow{2}{*}{Test}  & \# total   inst & 2,164   & 4,482   & 15,509\\
                       & \# pos inst     & 330     & 1,045   & 3,123\\\hline
\end{tabular}%
}
\caption{\small Statistics of used datasets.}
\label{table:dataset}
\vspace{-2mm}
\end{table}

\subsubsection{Baseline Methods}
We compare the proposed \our with three categories of methods including normal RE models, DSRE methods, and zero-shot RE methods.
For normal RE models, we consider two representative models BiLSTM \cite{zhang2015bidirectional} and BERT$\rm_{Entity Mask}$ \cite{zhou2021improved}. They both use the position information of entity mentions. 
For DSRE methods, we consider two recent methods ARNOR \cite{jia2019arnor} and SENT \cite{ma2021sent}. ARNOR achieves the SOTA performance on NYT2, while SENT achieves the SOTA performance on NYT1.
For zero-shot RE methods, we consider the NLI-based RE method \cite{sainz2021label}, only using the human written general template for each relation.

\begin{table*}[t]
\small
\centering
\resizebox{\textwidth}{!}{%
\begin{tabular}{cccc|ccc|ccc|ccc}
\hline
\multirow{2}{*}{Method} & \multicolumn{3}{c}{NYT1-Dev} & \multicolumn{3}{c}{NYT1-Test} & \multicolumn{3}{c}{NYT2-Dev} & \multicolumn{3}{c}{NYT2-Test} \\ \cline{2-13} 
 & P & R & F1 & P & R & F1 & P & R & F1 & P & R & F1 \\ \hline
BiLSTM$^\dagger$ & 36.71 & 66.46 & 47.29 & 35.52 & 67.41 & 46.53 & 41.46 & 70.17 & 52.12 & 44.12 & 71.12 & 54.45 \\ 
BERT$\rm_{Entity Mask}$ & 44.71 & 77.81 & 58.38 & 45.05 & 77.27 & 56.92 & 45.04 & 78.93 & 57.36 & 48.75 & \underline{81.91} & 61.12 \\ \hline
ARNOR$^\dagger$ & 62.45 & 58.51 & 60.36 & 65.23 & 56.79 & 60.90 & \textbf{78.14} & 59.82 & 67.77 & \textbf{79.70} & 62.30 & 69.93 \\ 
SENT$^\dagger\rm _{BiLSTM+BERT}$ & 69.94 & 63.11 & 66.35 & \textbf{76.34} & 63.66 & 69.42 & - & - & - & - & - & - \\
SENT$\rm _{BiLSTM}$ & 58.97 & 47.63 & 52.70 & 58.53 & 45.76 & 51.36 & 55.69 & 56.83 & 56.26 & 56.89 & 58.47 & 57.67 \\ \hline
NLI$\rm _{DeBERTa}$-genr & \textbf{74.21} & 55.33 & 63.39 & \underline{70.20} & 52.12 & 59.83 & \underline{73.43} & 51.70 & 60.68 & \underline{75.24} & 51.77 & 61.34 \\ \hline
DSRE-NLI$\rm_{IPIN}$ & \underline{71.93} & \underline{79.59} & \textbf{75.56} & 68.03 & \underline{80.61} & \underline{73.79} & 69.26 & \underline{79.65} & \textbf{74.09} & 73.90 & 81.82 & \textbf{77.66} \\ 
DSRE-NLI$\rm_{NPIN}$ & 67.06 & \textbf{84.91} & \underline{74.94} & 68.80 & \textbf{
84.85} & \textbf{75.98} & 66.23 & \textbf{82.63} & \underline{73.53} & 68.59 & \textbf{84.40} & \underline{75.68} \\ \hline
\end{tabular}
}
\caption{\small Results are in \%, where the bests are in bold, and the runner-ups are underlined. $\dagger$ cites results from referenced papers.}
\label{table:results}
\end{table*}
\subsubsection{\our Setups}

\noindent\textbf{SARV.}
One of the authors constructs one general template for each relation without reading any example sentences from the corpus. 
We initially choose the top 10\% most frequent mined patterns and only keep those that consist of less than 10 tokens and at least one non-stop-word token. For computational efficiency, we retain at most 50 patterns for each relation for the following pattern grouping. After pattern grouping, patterns with frequency of at least 10 are eligible for manual screening. One author is presented with a pattern and one example sentence at a time, and is asked if this pattern can induce the target relation. 

\noindent\textbf{NLI Model.}
We use the pretrained DeBERTa v2 model \cite{he2020deberta} to implement the NLI-based RE method, and set the entailment probability threshold $\tau$ to 0.95, which is empirically suggested by \citet{sainz2021label} based on their study on a different dataset. This setting is used for both pattern grouping and IS.

\noindent\textbf{Multi-Class RE Model.}
We use BERT$\rm_{Entity Mask}$ model from \cite{zhou2021improved}.

\subsubsection{Main Results}
We summarize the comparison results in Table~\ref{table:results} on dev and test sets of NYT1 and NYT2, respectively. Note that DSRE-NLI does not use the dev set to tune any hyperparameter. The first category of baseline methods treat the distant annotations as ground truth labels, and the results show that they suffer from low precision. It validates that for the distantly annotated data, high false positive rate is the major issue. With the denoising process, the DSRE methods clearly improve the precision, and achieve significantly higher F1 scores. The zero-shot method obtains high precision, either the best or the runner-up among all methods, but suffer from low recall. It implies that the general template for each relation has considerably low semantic coverage. 

The results clearly show that the proposed \our framework significantly outperforms previous state-of-the-art methods. DSRE-NLI$\rm_{NPIN}$ outperforms on NYT1 test set by a margin of 6.56\% comparing with the best baseline (SENT$\rm _{BiLSTM+BERT}$), and DSRE-NLI$\rm_{IPIN}$ outperforms on NYT2 test set by a margin of 7.73\% comparing with the best baseline (ARNOR). Both consolidation strategies can improve the prediction performance, achieving the best and runner-up overall performance. Comparing the two strategies, DSRE-NLI$\rm_{NPIN}$ consistently achieve higher recall, since DSRE-NLI$\rm_{NPIN}$ obtains bigger and possibly noisier training data for positive classes.


\subsubsection{Ablation Study}\label{sec:abla}
We conduct ablation studies to investigate the contributions of DS, IS, and pattern mining and grouping to the overall DSRE performance. Table \ref{table:abl} summarizes the results on combined dev and test data because DSRE-NLI does not use the dev set to tune any hyperparameter.

The first category of studies examines the importance of pattern mining and grouping for IS only. With the additional chosen patterns, the F1 score increases 11.57\% and 9.61\% on the two datasets, respectively, comparing with using general templates only. The results indicate that SARV increases the semantic diversity of the relation verbalization. The only additional manual effort is to select patterns from candidates, which requires much less effort than designing diverse templates from scratch. On average, the pattern mining method can find around 44 patterns per relation initially (Table \ref{table:savr}), but after pattern grouping, most patterns are merged, resulting in just around 5 patterns per relation for manual screening. The human annotator chose around 2.5 patterns from them. More details can be found in Tables \ref{table:nyt1sarv} and \ref{table:nyt1patt} in Appendix.

The second category of studies examines the impact of IS consolidation for the DSRE task. 
It is clear that using DS directly for training, the multi-class RE model performs poorly. Using IS (e.g., IPIN) with general template for data consolidation, we can see that the overall performance boosts significantly, gaining 11.57\% and 9.61\% of F1 scores on the two datasets, respectively. With selected patterns mined from the corpus, the performance further improves for 4.17\% and 1.34\%, respectively.  


\begin{table}[]
\centering
\resizebox{0.48\textwidth}{!}{%
\begin{tabular}{ll|l}
\hline
\multicolumn{1}{c}{\multirow{2}{*}{Method}} & \multicolumn{1}{c}{NYT1-(D+T)} & \multicolumn{1}{c}{NYT2-(D+T)} \\ \cline{2-3}
\multicolumn{1}{c}{}                        & \multicolumn{1}{c|}{F1}                 & \multicolumn{1}{c}{F1}                 \\\hline
NLI$\rm _{DeBERTa}$ & &                                                                                                      \\
\hspace{1mm}-genr                                        & 61.63                                  & 61.02                                  \\
\hspace{1mm}-genr + patt                                 & 73.20 (+11.57)                         & 70.63 (+9.61)                          \\\hline
DSRE-NLI & &                                                                                            \\
\hspace{1mm}-DS                                        & 57.65                                  & 59.28                                  \\
\hspace{1mm}-IPIN(DS, IS(genr))                              & 70.50 (+12.85)                         & 74.58 (+15.30)                         \\
\hspace{1mm}-IPIN(DS, IS(genr+patt))                      & \textbf{74.67} (+17.02)                         & \textbf{75.92} (+16.64)                        \\ \hline
\end{tabular}%
}
\caption{\small Results of two categories of ablation studies.}
\label{table:abl}
\end{table}

\begin{table}[]
\centering
\resizebox{0.48\textwidth}{!}{%
\begin{tabular}{cc|c|c}
\hline
Dataset & \# initial patt & \# patt after grouping & \# selected patt \\\hline
NYT1 & 45.6   & 5.2        & 2.5      \\\hline
NYT2 & 43     & 5         & 2.45     \\\hline
\end{tabular}%
}
\caption{\small Average of the number of patterns over all relations.}
\label{table:savr}
\vspace{-3mm}
\end{table}

\begin{table}[t]
\centering
\resizebox{0.48\textwidth}{!}{%
\begin{tabular}{ccc|cc}
\hline
\multirow{2}{*}{Pattern}                    & \multicolumn{2}{c}{After SARV} & \multicolumn{2}{c}{Before SARV} \\\cline{2-5}
                                            & Rank           & Freq          & Rank           & Freq           \\\hline
\textsf{\{subj\} , president of \{obj\}} (\cmark)             & 1              & 192           & 16             & 7              \\
\textsf{\{obj\} , Prime Minister \{subj\}} (\cmark)           & 2              & 89            & 19             & 7              \\
\textsf{\{subj\} , who led \{obj\}} (\xmark)                  & 3              & 87            & 27             & 4              \\
\textsf{\{obj\} 's foreign minister , \{subj\}} (\cmark)      & 4              & 48            & 3              & 45             \\
\textsf{\{obj\} 's former prime minister , \{subj\}} (\cmark) & 5              & 23            & 12             & 9              \\
\textsf{\{obj\} President \{subj\}} (\cmark)                  & 6              & 21            & 35             & 4              \\
\textsf{\{obj\} named \{subj\}} (\xmark)                      & 7              & 20            & 49             & 3              \\
\textsf{\{obj\} 's acting prime minister , \{subj\}} (\cmark) & 8              & 19            & 5              & 19            \\\hline
\end{tabular}%
}
\caption{\small Patterns of \texttt{nationality} generated by SARV.}
\label{table:nat_patt}
\end{table}

\begin{table}[t]
\centering
\resizebox{0.48\textwidth}{!}{%
\begin{tabular}{cc|ccc}
\hline
\multicolumn{2}{c}{NYT1-Train}        & \multicolumn{3}{c}{NYT1-(D+T)} \\\hline
Training data                   & \# instance & P         & R         & F1        \\\hline
DS                     & 8,355       & 58.46     & \textbf{92.68}     & 71.70     \\
IPIN(DS, IS(genr+patt)) & 2,850       & \textbf{73.33}     & 80.49     & 76.74     \\
NPIN(DS, IS(genr+patt)) & 3,591       & 72.92     & 85.37     & \textbf{78.65}    \\\hline
\end{tabular}%
}
\caption{\small Results of \texttt{nationality} in various settings.}
\label{table:nat_perf}
\vspace{-3mm}
\end{table}


\begin{table}[t]
\centering
\resizebox{0.45\textwidth}{!}{%
\begin{tabular}{ccc|cc}
\hline
Training data           & \# TP  & \# FP     & \# TN  & \# FN    \\\hline
TACREV                 & 13,012 & -         & 55,112 & -        \\\hline
TACREV-S               & 10,256 & 9,838     & 49,090 & 2,756    \\
                      &        & (48.96\%) &        & (5.32\%) \\\hline
IPIN(S-DS, IS)  & 4,895  & 898       & 47,519 & 1,499    \\
                      &        & (15.50\%) &        & (3.06\%) \\\hline
NPIN(S-DS, IS)  & 6,419  & 3,331     & 47,519 & 1,499    \\
                      &        & (34.16\%) &        & (3.06\%)\\\hline
\end{tabular}%
}
\caption{\small Training data quality in different settings.}
\label{table:deno}
\end{table}


\begin{table}[]
\centering
\resizebox{0.45\textwidth}{!}{%
\begin{tabular}{lccc}
\hline
\multirow{2}{*}{Method}   & \multicolumn{3}{c}{TACREV-Test} \\\cline{2-4}
                          & P         & R        & F1       \\\hline
BERT$\rm_{EntityMask}$-S-DS                     & 53.75     & \textbf{69.48}    & 60.62    \\\hline
SENT$\rm_{BiLSTM}$-S-DS     & 64.69     & 37.78    & 47.71    \\\hline
NLI$\rm_{DeBERTa}$                      &           &          &          \\
\hspace{1mm}-genr                     & \underline{83.82}     & 39.48    & 53.68    \\
\hspace{1mm}-temp$^\dagger$                     & 80.02     & 49.25    & 60.97\\
\hspace{1mm}-genr + patt              & 78.75     & 49.95    & 61.13    \\\hline
DSRE-NLI                  &           &          &          \\
\hspace{1mm}-IPIN(S-DS, IS(gt+patt)) & \textbf{84.59}     & 47.97    & \underline{61.22}    \\
\hspace{1mm}-NPIN(S-DS, IS(gt+patt)) & 75.95     & \underline{55.62}    & \textbf{64.21}   \\\hline
\end{tabular}%
}
\caption{\small Results on TACREV test set.}
\label{table:tacrev}
\vspace{-3mm}
\end{table}

\subsubsection{Case Study}
We use \texttt{nationality} from NYT1 dataset as an example relation to illustrate the process of \our. 
Table \ref{table:nat_patt} illustrates the relation patterns generated by SARV. The patterns are sorted by the frequency after pattern grouping, and the check marks after the patterns indicate if the pattern is selected. We can see that the ranks and the frequencies before and after pattern grouping are very different, indicating the grouping can merge semantic similar patterns. It is interesting to see that most patterns imply that the person is the leader of the country, which indeed implies the person's nationality. Human may find it hard to construct such patterns without reading the corpus.  

The relation extraction performance on \texttt{nationality} is shown in Table \ref{table:nat_perf}. We can see that both IPIN and NPIN strategies can significantly reduce the training instances of this relation, but the performance are both improved on dev and test sets, indicating that the removed instances are likely to be noise.  


\subsection{Experiments on Simulated Distantly Annotated Datasets}\label{sec:simexp}
Since there is no human annotation for the training data of NYTs, we cannot evaluate directly for the effectiveness of \our in improving training data quality. Therefore, we simulate the distant annotations on TACREV dataset (TACRED with revised dev ant test sets) \cite{zhang2017position, alt2020tacred}, a human-annotated relation extraction dataset, to quantitatively evaluate \our on improving training data. The statistics of the original TACREV dataset can be found in Table \ref{table:dataset}.

\subsubsection{Simulation Process}
To simulate the effect of distant annotation, we introduce both false positive (FP) errors and false negative (FN) errors in training data. We manipulate the original training instances from the perspective of entity pairs. To add FN errors, we first define long-tail entity pairs: if an entity pair is mentioned by $n$ sentences and $n$ is less than a threshold, then it is of long-tail. We relabel the instances mentioning long-tail entity pairs as NA to simulate the effect of limited coverage of knowledge bases. We empirically set the threshold so that FN rate is about 5\% based on our estimation from NYT datasets. To add FP errors, we follow the distant annotation process: if an entity pair participates in a relation, then all sentences in the training corpus mentioning the entity pair are labeled as positive instances of that relation. 
The statistics for the simulated training dataset, TACREV-S, can be found in Table \ref{table:deno}.


\subsubsection{Main Results}
Table \ref{table:deno} also shows the training data statistics obtained after IPIN(S-DS, IS(genr+patt)) and NPIN(S-DS, IS(genr+patt)) processes. It is clear that both strategies reduce FP and FN rate significantly, especially IPIN strategy. 

Table \ref{table:tacrev} demonstrates that the RE performance on the test set of TACREV using different TACREV-S training data. Similar to NYT datasets, the normal RE model encounters significant performance drop. SENT$\rm _{BiLSTM}$ improves the precision but experiences significantly low recall, indicating that this method may be too aggressive in denoising. In the zero-shot category, temp$^\dagger$ uses all human generated templates provided by \citet{sainz2021label}. We also compare with our designed general templates and general templates with pattern enrichment. 
We can see that templates can play an important role, and our SARV is comparable with the human designed templates. DSRE-NLI$\rm_{NPIN}$ achieves the best results on this dataset.

\section{Conclusion}
In this work, we propose a novel \our framework considering indirect supervision given by pretrained NLI models in DSRE tasks. We also design a novel SARV method to reduce the template design effort required by NLI-based zero-shot RE methods. With two simple and effective data consolidation strategies, the quality of training data is substantially improved. Extensive experiments demonstrate that the proposed framework significantly improves the SOTA performance on distantly supervised RE benchmark datasets.

\newpage
\bibliography{custom}
\bibliographystyle{acl_natbib}
\newpage
\clearpage
 
\section*{\LARGE Appendix} 
\setcounter{section}{0}
\section{Instance Generation for NYTs} \label{apx:inst}
The statistics of NYT1 dataset in \citet{jia2019arnor} is slightly different from ours in Table \ref{table:dataset} since the original NYT1 instances do not contain the position information of entity mentions. For example, the sentence `\textbf{Israel} claims all of \textbf{Jerusalem} as its capital, and government officials often describe Maale Adumim as part of greater \textbf{Jerusalem} that will be part of \textbf{Israel} in any future peace agreement.' mentions the subject \textbf{Israel} and the object \textbf{Jerusalem} multiple times. The original instance of this sentence in NYT1 dataset only claims that the subject and the object has a relation of \texttt{capital} without specifically pointing out which pair of mentions express the relation. 

In order to generate the position information of the subject and the object mentions, which is required by our framework and other baseline methods, we regenerate the instances. 

For the train set, we enumerate all entity pair combinations and generate multiple instances from such a sentence, where each instance has the same relation label. It is no doubt that this enumeration will bring more noisy instances into the training data. However, it is the common practice in the distant annotation process. Additionally, from the perspective of learning, if a DSRE method can deal with such noisy instances, then we believe that it has the capability to accurately capture and process the semantic features from local context fragments, which is truly desired in real applications. 
For the test set, we manually check on those sentences with multiple mentions of a specific entity and decide which pair of entity mentions expresses the relation. 

We do the same instance generation for NYT2 dataset. We will release the two newly constructed NYT datasets with the position information of entity mentions  for subsequent studies.

\section{Additional Experiments}
\subsection{Fine-tune NLI Models with Distantly Annotated Training Data}
Since fine-tuning the NLI model using training data demonstrates significant improvement for RE tasks in previous work \cite{sainz2021label}, we also examine if the distant annotations can be used directly to fine-tune the NLI model instead of training the multi-class RE model. we follow the fine-tuning process in \citet{sainz2021label}, and fine-tune the NLI model using distant annotations and consolidated annotations using IPIN strategy, respectively. The results are summarized in Table \ref{table:ft}. The fine-tuning with consolidated annotations using IPIN is better than no fine-tuning, but the DSRE-NLI$\rm_{IPIN}$ with the multi-class RE model still performs significantly better (see Table \ref{table:abl}). We conclude that using fine-tuned NLI model in replacement of the multi-class RE model is not recommended in DSRE tasks due to the noise in training data used for fine-tuning. 

\begin{table}[h]
\centering
\resizebox{0.48\textwidth}{!}{%
\begin{tabular}{lccc}
\hline
\multicolumn{1}{c}{\multirow{2}{*}{Method}} & \multicolumn{3}{c}{NYT2-(D+T)} \\\cline{2-4}
\multicolumn{1}{c}{}                        & P         & R         & F1        \\\hline
NLI$\rm_{DeBERTa}$-genr (no FT)                                      & 74.36     & 51.73     & 61.02     \\
\hspace{1mm}-FT on   DS                                 & 45.32     & 78.05     & 57.35     \\
\hspace{1mm}-FT on   IPIN          & 53.64     & 83.55     & 65.34    \\\hline
\end{tabular}%
}
\caption{\small Fine-tuning the NLI model with different training data.}
\label{table:ft}
\end{table}

\begin{table}[h]
\centering
\resizebox{0.48\textwidth}{!}{%
\begin{tabular}{cc|c|c}
\hline
 & NYT1-Test  & NYT2-Test  & TACREV-S-Test \\\hline
EM         & 75.98         & 75.68         & 64.21         \\
                      & (68.80/84.85) & (68.59/84.40) & (75.95/55.62) \\\hline
TEM   & 69.19         & 69.94         & 64.70         \\
                      & (57.99/85.76) & (57.81/88.52) & (76.29/56.16)\\\hline
\end{tabular}%
}
\caption{\small Comparison of entity representations for DSRE on test sets: F1 score (Precision/Recall) (in \%).}
\label{table:rep}
\end{table}

\begin{table*}[p]
\centering
\resizebox{1\textwidth}{!}{%
\begin{tabular}{l|c|c|c}
\hline
Relation                                  & \# initial patt & \# patt after grouping & \# selected patt \\\hline
/location/location/contains               & 50              & 10                     & 1                \\
/people/person/nationality                & 50              & 8                      & 6                \\
/location/country/capital                 & 50              & 2                      & 0                \\
/people/person/place\_lived               & 50              & 12                     & 8                \\
/business/person/company                  & 50              & 11                     & 9                \\
/location/neighborhood/neighborhood\_of   & 50              & 1                      & 0                \\
/people/person/place\_of\_birth           & 50              & 4                      & 0                \\
/people/deceased\_person/place\_of\_death & 50              & 0                      & 0                \\
/business/company/founders                & 37              & 4                      & 1                \\
/people/person/children                   & 19              & 0                      & 0               \\\hline
\end{tabular}%
}
\caption{Effectiveness of SAVR on NYT1 dataset.}
\label{table:nyt1sarv}
\end{table*}

\begin{table*}[p]
\centering
\resizebox{\textwidth}{!}{%
\begin{tabular}{l|l|l}
\hline
Relation                                     & Templates (genr   + patt)                   & NER type constraints                  \\\hline
\multirow{2}{*}{/location/location/contains} & \{obj\} is located in \{subj\}              & LOCATION:LOCATION                     \\
                                             & \{subj\} , including \{obj\}                & LOCATION:ORGANIZATION                 \\\hline
\multirow{7}{*}{/people/person/nationality}  & \{obj\} is the nationality of \{subj\}      & \multirow{7}{*}{PERSON:LOCATION}      \\
                                             & \{subj\} , president of \{obj\}             &                                       \\
                                             & \{obj\} , Prime Minister \{subj\}           &                                       \\
                                             & \{obj\} 's foreign minister , \{subj\}      &                                       \\
                                             & \{obj\} 's former prime minister , \{subj\} &                                       \\
                                             & \{obj\} President \{subj\}                  &                                       \\
                                             & \{obj\} 's acting prime minister , \{subj\} &                                       \\\hline
/location/country/capital                    & \{obj\} is the capital of \{subj\}          & LOCATION:LOCATION                     \\\hline
\multirow{9}{*}{/people/person/place\_lived} & \{subj\} lives in \{obj\}                   & \multirow{9}{*}{PERSON:LOCATION}      \\
                                             & \{subj\} , Republican from \{obj\}          &                                       \\
                                             & \{subj\} , Democrat of \{obj\}              &                                       \\
                                             & \{obj\} Gov. \{subj\}                       &                                       \\
                                             & \{obj\} mayor , \{subj\}                    &                                       \\
                                             & \{subj\} , the former senator from \{obj\}  &                                       \\
                                             & \{subj\} , who was mayor when \{obj\}       &                                       \\
                                             & \{obj\} Coach \{subj\}                      &                                       \\
                                             & \{obj\} State Coach \{subj\}                &                                       \\\hline
\multirow{10}{*}{/business/person/company}   & \{subj\} works in \{obj\}                   & \multirow{10}{*}{PERSON:ORGANIZATION} \\
                                             & \{subj\} , the head of \{obj\}              &                                       \\
                                             & \{obj\} 's chairman , \{subj\}              &                                       \\
                                             & \{obj\} 's president , \{subj\}             &                                       \\
                                             & \{subj\} chief executive , \{obj\}          &                                       \\
                                             & \{obj\} secretary general , \{subj\}        &                                       \\
                                             & \{subj\} , a professor at \{obj\}           &                                       \\
                                             & \{subj\} , founder of \{obj\}               &                                       \\
                                             & \{obj\} newsman \{subj\}                    &                                       \\
                                             & \{subj\} , the co-founder of \{obj\}        &                                       \\\hline
/location/neighborhood/neighborhood\_of      & \{subj\} is in the neighborhood of \{obj\}  & LOCATION:LOCATION                     \\\hline
/people/person/place\_of\_birth              & \{subj\} was born in \{obj\}                & PERSON:LOCATION                       \\\hline
/people/deceased\_person/place\_of\_death    & \{subj\} died in \{obj\}                    & PERSON:LOCATION                       \\\hline
\multirow{2}{*}{/business/company/founders}  & \{subj\} was founded by \{obj\}             & \multirow{2}{*}{ORGANIZATION:PERSON}  \\
                                             & \{subj\} co-founder \{obj\}                 &                                       \\\hline
/people/person/children                      & \{subj\} is the parent of \{obj\}           & PERSON:PERSON                      \\\hline  
\end{tabular}%
}
\caption{Templates generated and selected by SARV for NYT1 dataset. The first template is the general template in each relation, and the other templates are mined patterns.}
\label{table:nyt1patt}
\end{table*}

\subsection{Entity Representation in BERT}
As explained in Section \ref{sec:MRE}, we adopt the architecture proposed by \citet{zhou2021improved} using Entity Mask (EM), where the entities are replaced with the entity NER types. A better entity representation method found in \citet{zhou2021improved} is Typed Entity Marker (TEM), where entity type markers are added surrounding entities.
We also empirically compare EM and TEM on the distantly annotated datasets. The results are summarized in Table \ref{table:rep}. The results confirmed our intuition.  Without information for the entity surface names, the model learns from the context of the sentences instead of reliance on prior knowledge, and thus more robust to label noise in the DS setting.

\subsection{SARV Generated Templates for NYT1}

This section describes the templates generated and selected by the SARV method in the NYT1 experiments. In Table \ref{table:nyt1sarv}, we present the number of patterns before and after pattern grouping, and the number of patterns selected by the human annotator for each relation type. We can see that the automated pattern mining and grouping is effective in reducing candidate patterns, and thus reducing the manual effort. Table \ref{table:nyt1patt} demonstrates all templates used in the NYT1 experiments. The first template for each relation is the human designed general template. We can see that the selected templates from mined patterns tend to be more specific. For example, the patterns for relation \texttt{/business/person/company} express various job titles including chairman, chief executive, and professor, which may be hard for human annotators to generate without prior knowledge of the target corpus.

\section{Other Setups}
We run all methods using one Tesla V100S GPU (32G). We train DSRE-NLI for 2 epochs on both NYT1 and NYT2 training variations, 5 epochs on TACREV-S training variations.
\end{document}